\newcommand{\xBc}{\langle}
\newcommand{\xBe}{\rangle}

\newcommand{\xbS}{\Sigma}

\newcommand{\xba}{\alpha}

\newcommand{\xbe}{\in}
\newcommand{\xbf}{\phi}

\newcommand{\xbi}{\iota}

\newcommand{\xbm}{\mu}
\newcommand{\xbn}{\nu}

\newcommand{\xbq}{\psi}
\newcommand{\xbr}{\rho}

\newcommand{\xCN}{\neg}

\newcommand{\xCf}{\hspace{0.1em}}

\newcommand{\xcA}{\forall}

\newcommand{\xcE}{\exists}

\newcommand{\xcg}{\geq}

\newcommand{\xck}{\leq}

\newcommand{\xco}{\vee}
\newcommand{\xcp}{\rightarrow}

\newcommand{\xDH}{\item }

\newcommand{\xdr}{{\cal R}}

\newcommand{\xEI}{\begin{itemize}}
\newcommand{\xEJ}{\end{itemize}}

\newcommand{\xEh}{\begin{enumerate}}

\newcommand{\xEj}{\end{enumerate}}

\newcommand{\xEn}{\begin{description}}
\newcommand{\xEp}{\end{description}}

\newcommand{\xeA}{\nabla}

\newcommand{\bl}{\begin{lemma} \rm}
\newcommand{\el}{\end{lemma}}
\newcommand{\br}{\begin{remark} \rm}
\newcommand{\er}{\end{remark}}
\newcommand{\be}{\begin{example} \rm}
\newcommand{\ee}{\end{example}}
\newcommand{\bco}{\begin{corollary} \rm}
\newcommand{\eco}{\end{corollary}}
\newcommand{\bc}{\begin{claim} \rm}
\newcommand{\ec}{\end{claim}}
\newcommand{\bfa}{\begin{fact} \rm}
\newcommand{\efa}{\end{fact}}
\newcommand{\bp}{\begin{proposition} \rm}
\newcommand{\ep}{\end{proposition}}
\newcommand{\bd}{\begin{definition} \rm}
\newcommand{\ed}{\end{definition}}
\newcommand{\bcs}{\begin{construction} \rm}
\newcommand{\ecs}{\end{construction}}
\newcommand{\bcd}{\begin{condition} \rm}
\newcommand{\ecd}{\end{condition}}
\newcommand{\bt}{\begin{theorem} \rm}
\newcommand{\et}{\end{theorem}}
\newcommand{\bn}{\begin{notation} \rm}
\newcommand{\en}{\end{notation}}
\newcommand{\bfi}{\begin{bild} \rm}
\newcommand{\efi}{\end{bild}}
\newcommand{\bsta}{\begin{statement} \rm}
\newcommand{\esta}{\end{statement}}
\newcommand{\bcom}{\begin{comment} \rm}
\newcommand{\ecom}{\end{comment}}
\newcommand{\bdia}{\begin{diagram} \rm}
\newcommand{\edia}{\end{diagram}}

\newcommand{\bfc}{\begin{figure}[htb] \begin{center}}
\newcommand{\efc}{\end{center} \end{figure}}

\documentclass{article}

\usepackage{amssymb,latexsym,epic,eepic,rotating}

\title{A Reliability Theory of Truth
\thanks{File tru
}
}

\author{Karl Schlechta
\thanks{
schcsg@gmail.com - https://sites.google.com/site/schlechtakarl/ -
Koppeweg 24, D-97833 Frammersbach, Germany}
\thanks{
Retired, formerly: Aix-Marseille Universit\'{e}, CNRS, LIF UMR 7279, F-13000
Marseille, France
}
}

\begin{document}

\newtheorem{lemma}{Lemma}[section]
\newtheorem{theorem}[lemma]{Theorem}
\newtheorem{proposition}[lemma]{Proposition}
\newtheorem{corollary}[lemma]{Corollary}
\newtheorem{claim}[lemma]{Claim}
\newtheorem{fact}[lemma]{Fact}
\newtheorem{remark}[lemma]{Remark}
\newtheorem{definition}{Definition}[section]
\newtheorem{construction}{Construction}[section]
\newtheorem{condition}{Condition}[section]
\newtheorem{example}{Example}[section]
\newtheorem{notation}{Notation}[section]
\newtheorem{bild}{Figure}[section]
\newtheorem{comment}{Comment}[section]
\newtheorem{statement}{Statement}[section]
\newtheorem{diagram}{Diagram}[section]

\renewcommand{\labelenumi}
  {(\arabic{enumi})}
\renewcommand{\labelenumii}
  {(\arabic{enumi}.\arabic{enumii})}
\renewcommand{\labelenumiii}
  {(\arabic{enumi}.\arabic{enumii}.\arabic{enumiii})}
\renewcommand{\labelenumiv}
  {(\arabic{enumi}.\arabic{enumii}.\arabic{enumiii}.\arabic{enumiv})}

\maketitle

\setcounter{secnumdepth}{3}
\setcounter{tocdepth}{3}

\begin{abstract}

Our approach is basically a coherence approach, but we avoid the well-known
pitfalls of coherence theories of truth. Consistency is replaced by
reliability, which expresses support, and, in principle, every
theory (or agent, message) counts. At the same time, we do not require a
priviledged access to "reality". Consequently, we also modify the notions
of semantics and representation theorems.

A centerpiece of our approach is that
we attribute reliability also to agents, messages, etc., so an unreliable
source of information will be less important in future.

Our ideas can also be extended to value systems, and even actions, e.g.,
of animals.

Technically, but not in motivation, we are sometimes closer to systems of trust
and argmentation in computer science, than to philosophical truth theories.

\end{abstract}

% HAL: id: hal-01672320, pw: &#b561

% arXiv id 1801.01788, pw 3fjzk

\tableofcontents

% *** BEGIN LATEX SOURCE tru-1.tex ***
%
% Aus Karltex File: tru-1.ms
%
%
\section{
Introduction
}
\subsection{
The Coherence and Correspondence Theories of Truth
}

See  \cite{Sta17a} for an overview for the coherence theory,
and  \cite{Sta17b} for an overview for the correspondence theory.
The latter contains an extensive bibliography, and we refer the
reader there for more details on the correspondence theory.

We think that the criticisms of the coherence theory of truth are
peripheral,
but the criticism of the correspondence theory of truth is fundamental.

The criticism of the correspondence theory, that we have no direct access
to reality, and have to do with our limitations in observing and thinking,
seems fundamental to the author. The discussion whether there are some
``correct'' theories our brains are unable to formulate, is taken seriously
by physicists, likewise the discussion, whether e.g. Quarks are real, or
only helpful ``images'' to understand reality, was taken very seriously.
E.g. Gell Mann was longtime undecided about it, and people perhaps just
got used to them. We don't know what reality is, and it seems we will
never know.
See also discussions in Neurophilosophy,
 \cite{Sta17d} for a general introduction.

On the other side, two main criticisms of the coherence theory can be
easily countered, in our opinion.
See e.g.  \cite{Rus07} and  \cite{Tha07} for objections to
coherence theory.
Russell's objection, that $ \xbf $ and $ \xCN \xbf $
may both be consistent with a given theory, shows just that ``consistency''
is the wrong interpretation of ``coherence'', and it also leaves open the
question which logic we work in. The objection that the background
theory against which we check coherence is undefined, can be countered
with a
simple
argument: Everything. In ``reality'', of course, this is not the case.
If we have a difficult physical problem, we will not ask our baker, and
even if he has an opinion, we will not give it much consideration.
Sources of information are assessed, and only ``good'' sources (for the
problem at hand!) will be considered.
(Thus, we also avoid the postmodernist trap: there are standards of
``normal reasoning'' whose values have been shown in unbiased everyday
life, and against which standards of every society have to be compared.
No hope for the political crackpots here!)

Our approach will be a variant of the coherence theory,
related ideas were also expressed by  \cite{Hem35} and  \cite{Neu83}.

We can see our approach in the tradition of relinquishing absoluteness:
 \xEI
 \xDH
The introduction of axiom systems made truth relative to axioms.
 \xDH
Nonmonotonic reasoning allowed for exceptions.
 \xDH
Our approach treats uncertainty of information, and our potential
inability to know reality.
 \xEJ
\subsubsection{
A Short Comparison
}

 \xEh

 \xDH
Our approach is not about discovery, only about evaluating information.

 \xDH
In contrast to many philosophical theories of truth,
we do not treat paradoxa, as done e.g. in
 \cite{Kri75} or  \cite{BS17},
we assume statements to be ``naive'' and free from semantic problems.

We do treat cycles too, but they are simpler, and we take care not to go
through them repeatedly. In addition, our structures are assumed to be
finite.

 \xDH
On the philosophical side, we are probably closest to the discourse
theory of the Frankfurt School,
in particular to the work by J. Habermas and K. O. Apel
(as we discovered by chance!),
see e.g.
 \cite{Wik18b},  \cite{Sta18b},  \cite{Hab73},
 \cite{Hab90},  \cite{Hab96},  \cite{Hab01},  \cite{Hab03}.

Importantly, they treat with the basically same methods problems of
truth and ethics, see our
Remark \ref{Remark Extensions} (page \pageref{Remark Extensions})  below.

We see three differences with their approach.
 \xEh
 \xDH
A minor difference: We also consider objects like thermometers as agents,
not only human beings, thus eliminating some of the subjectivity.
 \xDH
A major difference: We use feedback to modify reliability of agents and
messages. Thus, the $ \xcA $-quantifier over participating agents in the
Frankfurt
School is attenuated to those considered reliable.
 \xDH
Conversely, their discourse theory is, of course, much more developed than
our approach.
 \xEj

Thus, an integration of both approaches seems promising.

 \xDH
Articles on trust, like
 \cite{BBHLL10} or  \cite{BP12}, treat different, more subtle,
and perhaps less
fundamental, problems.
A detailed overview over trust systems is given in
 \cite{SS05}.

We concentrate on logics, cycles, and composition of values by
concatenation.
Still, our approach is in methods, but not in motivation, perhaps closer
to the
basic ideas of trust systems,
than to those of theories of truth, which often concentrate on paradoxa.

Articles on trust will often describe interesting ideas about details of
coding,
e.g.  \cite{BP12} describes how
to code a set of numerical values by an interval (or, equivalently, two
values).

 \xDH
Basic argumentation systems, see e.g.  \cite{Dun95},
will not distinguish between arguments of different quality.
Argumentation systems with preferences,
see e.g.  \cite{MP13}, may do so, but they do not seem to
propagate conflict and confirmation backwards to the source of arguments,
which
is an essential part of our approach. This backward
propagation also seems a core part of any truth theory
in our spirit. Such theories have to be able to learn
from past errors and successes.
 \xEj
\subsection{
Two Examples
}

Some examples may help to illustrate our ideas.

\be

$\hspace{0.01em}$

{\tiny (+++ Orig. No.:  Example Size +++)}

{\tiny LABEL: {Example Size}} \\[1mm]
\label{Example Size}

Suppose we are interested in the size of an object $X.$
We cannot access $X$ directly, and have to rely on witnesses.

Witness $ \xCf A$ had a meter, measured $X,$ and says $X$ is 120 cm long.
Unfortunately, $ \xCf A$ is known to be crackpot.

Thus, we limit our sources of information to reliable ones.

Witness $B$ had no meter, he measured using his thumb, and later
calculated
the length to be 90 cm.

Witness $C$ had a meter, but the meter was old and twisted, so not very
accurate. $C$ says that $X$ is 101 cm long.

By experience, $ \xCf C$'s method is superior to $ \xCf B$'s method.

This is all we know.

Based on this information, we say ``our best estimate is that $X$ is 101 cm
long''.

We do not doubt that there is some ``real'' length of $X,$ but this is
irrelevant, as we cannot know it. We have to do with what we know, but
are aware that additional information might lead us to revise our
estimate.

\ee

This story seems simple, but even much more complicated stories can be
solved by essentially the same, simple ideas, we think.

\br

$\hspace{0.01em}$

{\tiny (+++ Orig. No.:  Remark Extensions +++)}

{\tiny LABEL: {Remark Extensions}} \\[1mm]
\label{Remark Extensions}

The following extensions seem possible:
 \xEI
 \xDH Actions and animals
We can apply similar reasoning to actions. The action
of a monkey which sees a lion and climbs a tree to safety is ``true'', or,
better, adequate.
 \xDH Values
Values, obligations, ``natural laws'' are subjective. Still, some influences
are
known, and we can try to peel them off. Religion, politics, personal
history, influence our ideas about values. One can try to find the
``common'' and ``reasonable'' core of them. For instance, religious extremism
tends to produce ruthless value systems, so we might consider relious
extremists as less reliable about values.
 \xEJ

\er

We give another example.
Reliabilities will be denoted $ \xbr (.).$

\be

$\hspace{0.01em}$

{\tiny (+++ Orig. No.:  Example Temp +++)}

{\tiny LABEL: {Example Temp}} \\[1mm]
\label{Example Temp}

We have a meteorological station in Siberia. The thermometer is supposed
to be reliable, it automatically records the current temperature (with
time stamp etc.) reliably. $(\xbr =1$ below. This is ``reality'', which is
introduced
purely as a trick, to illustrate simple cases, we discard this later.)
Sometimes, we want to know the current temperature
immediately, we phone the human operator or operators and ask.
Unfortunately,
the line or lines is/are very noisy, and errors in transmission occurr.
This is the common part.
\xEn
 \xDH
Case 1:

The human operator is absolutely reliable. Later, we compare the
temperature
as transmitted by phone with the recorded temperature, and assign $ \xbr
(t)$
to the transmission $t.$

The values of the absolutely reliable thermometer etc. are unchanged.

 \xDH
Case 2:

We have two reliable human operators, they use different unreliable phone
lines. So we have transmissions $t$ and $t'.$ Assume that, e.g. based on
previous
experience,
we gave $t$ and $t' $ already initial values $ \xbr (t),$ $ \xbr (t').$

If $T$ and $T' $ agree ($T,$ $T' $ the temperatures received, without any
knowledge of
the ``real'' temperature!), we increase $ \xbr (t)$ and $ \xbr (t'),$ they
confirm each
other.

If $T$ and $T' $ disagree (without any knowledge of the ``real''
temperature!),
we decrease $ \xbr (t)$ and $ \xbr (t'),$ as they contradict each other,
based
on their initial values.

 \xDH
Case 3:

As it is so cold, the human operators often drink too much, so they are
not
reliable. They make mistakes, and the transmission is not reliable,
either.
(To simplify, we assume that mistakes do not cancel each other, e.g. the
operator
reads 10 degrees too much, and the line transmits 10 degrees too little,
so
the correct value is transmitted.)

\xEn
 \xDH
Case 3.1:

We consider only one human operator $h,$ and one transmission line $t,$
and compare
the value with the recorded temperature.
We first calculate the combined reliability of the chain $ \xCf ht,$ $
\xbr (ht).$

If the transmitted temperature agrees with the recorded value, we
increase the combined reliability $ \xbr (ht),$ and break this down to
increases of $ \xbr (h)$ and $ \xbr (t),$ according to their previous
values.

If they disagree, we first decrease the combined reliability $ \xbr (ht),$
and break
this down again to decreases of $ \xbr (h)$ and $ \xbr (t).$

 \xDH
Case 3.2:

We consider both operators and transmission lines, and compare the
received values.
Let $h,$ $h' $ be the human operators, $t,$ $t' $ the transmission lines.
We proceed as in Case 3.1, but first adjust both $ \xbr (ht)$ and $ \xbr
(h' t'),$
as in Case 2, and break the adjustments down to $ \xbr (h),$ $ \xbr (h'
),$
$ \xbr (t),$ $ \xbr (t')$ as in Case 3.1.
\xEp
\xEp
\section{
Concepts and Basic Ideas
}
\subsection{
Concepts
}
\subsubsection{
Information vs. Facts
}

{\tiny LABEL: {Section Information}} \\[1mm]
\label{Section Information}

\ee

 \xEh
 \xDH
Our idea is close to intuitionistic logic. We have to distinguish
information
and facts. We might be convinced (or informed with high reliability) that
$ \xbf \xco \xCN \xbf $ holds, but have information with low reliability
that $ \xbf $ and with
low reliability that $ \xCN \xbf.$ For instance, two witnesses saw a car,
one says
it was black, the other that it was not black, but both concede that
visibility was very bad, so they cannot be sure.

Thus, we should have $ \xbr (\xbf \xco \xbq)$ $ \xcg $ $max\{ \xbr (
\xbf), \xbr (\xbq)\},$ but
not necessarily equality.

The same example shows that it is not necessarily true that
$ \xbr (\xCN \xbf)=- \xbr (\xbf).$
 \xDH
The reliability order is independent of the usual truth value order
$false<true,$ this is obvious.
 \xDH
The reliability order is also independent of a normality order.
Thus, we might have reliable information that Tweety is an abnormal bird,
and unreliable information that $x$ is a normal bird.

Likewise, unreliability of $ \xbf \xcp \xbq $ is different from $ \xeA (
\xbf \xcp \xbq).$
In the first case, we may think that $ \xbf \xcp \xbq $ does always hold,
but
have no reliable information about this. In the second case,
we are certain that, in most cases, if $ \xbf $ holds, then so does $ \xbq
.$

Moreover, we may have uncertain information that $x$ is more normal than
Tweety.
 \xDH
Objects we talk about may have names. Two physicians may talk about the
same patient $x$ (they are sure about this), may be unsure about a
diagnosis,
and disagree about it.
 \xEj
\subsubsection{
Semantics and Representation Theorems
}

We have to distinguish semantics on the level of details (message,
reliability,
inertia, etc., agents which send and receive messages, process them,
etc.), and
semantics on the abstract level. The latter does not exist,
this would presume that there is some ``reality'', a notion we try to avoid.

But we can distinguish statements above a certain level of reliability,
this
is ``relative truth''. Recall that $ \xbf \xco \xbq $ might be more reliable
than the
known reliabilities of $ \xbf $ and
$ \xbq $ separately, so, depending on the level of reliability we chose,
formulas
may be analysed, or not. (Likewise for $ \xcE x \xbf,$ etc.)

Representation cannot be relative to a semantics, but only relative to
formulas above a certain level of reliability. Of course, this is
basically
the same as classical representation, but without the philosophical
overhead. Here, it is just about describing a certain set of formulas in a
perhaps concise way.

In preferential structures,
the normality relation may also have uncertain reliabilty (see above), so
we may
reason only with normality above a certain threshold.

Moreover, we may have static representation (about what holds beyond a
certain
level of reliability), and dynamic representation (about how values
develop
under new input).
\subsubsection{
Coding
}

The details of coding are not important here, we indicate some
problems and possible solutions.

 \xEI

 \xDH Inertia

If some value is supported by many sources, it should be more stable under
challenges than a value supported by only a few sources. We may code this
by
``inertia''.

 \xDH Loops

We want to avoid that person $x$ thinks highly about person $y,$ and vice
versa,
but they just defend each other. Such loops should be detected, and it
should be avoided to go in circles, enforcing each other.
For such reasons, we may code the path of messages, and their
consequences,
and use message $ID$'s.
Thus, every agent may detect if this message chain has been going
past him already.

On the other hand, if agent $x$ sends a wrong message, this should fall
back
on him, so this kind of cycle is a good one. Sending the whole chain
along allows to distinguish both cases.

(We do not pretend that this is how the brain works, it is just some way
to
achieve the aim.)

We illustrate this with an example

\be

$\hspace{0.01em}$

{\tiny (+++ Orig. No.:  Example Cycles +++)}

{\tiny LABEL: {Example Cycles}} \\[1mm]
\label{Example Cycles}

 \xEI
 \xDH
(a) We have an agent $ \xCf A,$ considered reliable, $ \xCf A$ sends a
message $M$ to agent $B$
that agent $A'' $ is reliable. $M$ is considered reliable (e.g. little
noise), too.
$B$ is considered reliable, and sends a message $M' $ to $A',$ that $A''
$ is reliable.
$M' $ is again considered reliable. $A' $ has past information that $A'' $
is indeed
reliable. So $A' $ answers $M',$ that $M' $ seems correct, and $M' $ has
increased
reliablity, going back to $B,$ $M,$ and $ \xCf A.$ But we should stop
here, and
not send $M$ again with increased reliability. Sending the history avoids
this.
 \xDH
(b) Agent $ \xCf A$ sends a message $M$ to $B$ that agent $A' $ is
reliable. $B$ considers
$ \xCf A$ and $A' $ somewhat reliable (e.g. by default), and increases the
reliability of
$A'.$ Now, $A' $ sends a message $M' $ to $B$ (with increased
reliability!) that $ \xCf A$
is reliable. So $B$ increases the reliability of $ \xCf A.$ This might
still be
acceptable, but we should not send $M$ again with increased reliability.
(We assume all messages are free from noise, reliable.)
Again, the history of the messages can detect such cycles.
 \xEJ

\ee

 \xEJ
\subsection{
Agents, Messages, and Reliabilities
}

\bd

$\hspace{0.01em}$

{\tiny (+++ Orig. No.:  Definition Agents +++)}

{\tiny LABEL: {Definition Agents}} \\[1mm]
\label{Definition Agents}

 \xEh
 \xDH
We define as usual $[-1,1]:=\{r \xbe \xdr:$ $-1 \xck r \xck 1\},$ etc.

 \xDH
We have agents $ \xCf A,$ $A' $ etc. and messages, $M,$ $M' $ etc.

Agents may be people, devices like thermometers, transmission lines,
theories, etc.

 \xDH
If agent $ \xCf A$ sends a message $M$ to agent $A',$ $ \xCf A$ will be
called the source of $M,$
$A' $ its destination.

 \xDH
Agents and messages have values of reliability, $ \xbr (A),$ $ \xbr (M),$
etc.
Sometimes, it is more adequate to see reliability as degree of competence,
for instance for moral questions.

 \xDH
Messages may be numbers, e.g. a temperature, but also opinions about
something, e.g. ``the earth is flat'',
also about the reliability of agents and messages.
Messages may also be moral judgements.
If a message is a moral judgement, then moral competence of the source of
the message is important. If the source is the constitution of a country,
the competence will probably be considered high, etc.
In case of theories, the messages may be consequences of this theory, etc.

 \xDH
A human agent may be a good chemist, but a poor mathematician, so his
reliability varies with the subject. We neglect this here, and treat this
agent as two diffent agents, $A$-Chemist, $A$-Mathematician, etc.
Likewise, a thermometer may be reliable between 0 and 30 $C,$ but less
reliable
below 0. Again, we may describe them as two different thermometers.

 \xDH Agents may also have doubts about the reliability of their own
messages,
a human agent may doubt its competence, a thermometer if the value is
out of its intended range.

 \xDH
 \xEh

 \xDH Version 1

Reliabilities are just values in $[-1,1],$ and ordered in the usual way.
$ \xbr =1$ is maximal reliability, $ \xbr =-1$ is maximal uncertainty.

 \xDH Version 2

Partial orders:

Reliabilities (of agents or messages) will be multisets of the form
$\{r_{i}a_{i}:i \xbe I\}.$ $r_{i}$ will be a real value between -1 and
$+1,$ $a_{i}$ should be
seen as a ``dimension''.

This allows for easy adjustment, e.g. ageing over time, shifting
importance,
etc., as we will shortly detail now:
 \xEI
 \xDH the real values allow arbitrarily fine adjustments, it is not just
$\{-1,0,1\},$
 \xDH the dimensions allow to treat various aspects in different ways,
 \xDH for instance, we can introduce new agents with a totally ``clean
slate'',
0 in every dimension, or preset some dimensions, but not others,
 \xDH the uniform treatment of all dimensions in
Section \ref{Section Combinations} (page \pageref{Section Combinations})  is not
necessary, we can treat different dimensions differently, e.g., conflicts
between two agents in dimension $a_{i}$ need not touch dimension $a_{j},$
etc.
 \xEJ
We have arbitrarily many dimensions, with possibly different meaning and
treatment, and within each dimension arbitrarily many values. This is not
a total order, but within each dimension, it is.

 \xEj

 \xEj
\subsection{
Basic Data Structure
}

\ed

 \xEh
 \xDH Values (the value of an agent is his reliability)
 \xEh
 \xDH short version

$ \xBc $ value, reliability (of value), inertia (of value) $ \xBe $

 \xDH long version

$ \xBc $ value, reliability (of value), history (of value) $ \xBe $

(history: past pairs $ \xBc $ value, reliability $ \xBe $)
 \xEj

 \xDH messages

 \xEh
 \xDH short version

$ \xBc $ value, reliability (of value) $ \xBe $

 \xDH long version

$ \xBc $ value, reliability (of value), history $ \xBe $

(history: chain of messages with source, destination,
and pairs $ \xBc $ value, reliability $ \xBe $ of messages)

In the long version, we may give each message chain an ID, so it can
easily
be identified, and distinguished from other messages. The ID is given at
the
start, and passed on to subsequent messages. (These are details of coding
the basic ideas.)

 \xEj
 \xEj
\subsubsection{
Comments
}

The long versions give the complete relevant history. Inertia may, e.g.,
code the number of messages already gone into the
$ \xBc value,reliability \xBe $ pair.

The history of messages allows to detect loops, directions of messages,
etc.
A messages chain may come back to a prior node (agent), but should not go
again in the same direction. This eliminates unwanted feedback, see
Example \ref{Example Cycles} (page \pageref{Example Cycles}).
The history may need more information than just outlined, our
emphasis is more on principles, than on details of coding.
\subsection{
Processing
}

As our discussion is mostly conceptual, it suffices to indicate problems
and solutions without going into too much detail.
In addition, we think that precise values may depend on the domain treated
(more or less caution), and their importance should be dampened by a
suitable overall algorithm.

 \xEh
 \xDH AND

When agent $ \xCf A$ with reliability $ \xbr (A)$ sends a message $M$ with
reliability
$ \xbr (M),$ then the combined reliability should be at most $min\{ \xbr
(A), \xbr (M)\}.$

Note that Modus Ponens is a form of AND, so it should be treated the same
way.

 \xDH OR

When two messages $M,$ $M' $ with the same value have reliability $ \xbr
(M)$ and
$ \xbr (M')$ respectively, then the value should have reliability at
least
$max\{ \xbr (M), \xbr (M')\}.$
See Section \ref{Section Information} (page \pageref{Section Information}).

 \xDH NOT

This was also discussed above in
Section \ref{Section Information} (page \pageref{Section Information}).

 \xEj
\section{
Details - an Example
}

{\tiny LABEL: {Section Details}} \\[1mm]
\label{Section Details}

In this section, we work with a total order. We may, however, extend the
approach to sequences of total orders (and thus to the partial order
idea) by an elementwise treatment. In this case, we may have to initialize
new dimensions to some default value:
If the operations work on more than one reliability, any dimension $ \xCf
a$ which
is present in some, but not all reliabilities, will be added where
necessary with the default value.

\bd

$\hspace{0.01em}$

{\tiny (+++ Orig. No.:  Definition Operations +++)}

{\tiny LABEL: {Definition Operations}} \\[1mm]
\label{Definition Operations}

 \xEh
 \xDH
If $r,r' \xbe [0,1],$ and $r+r' =1,$ we define the weighted mean of $ \xbr
$ and $ \xbr' $ as
$r* \xbr +r' * \xbr'.$ (Similarly for more than two $ \xbr $'s.)
 \xEj

For the partial order idea:

 \xEh
 \xDH
The average reliability of $ \xbr $ is defined as $av(\xbr):= \frac{
\xbS \{ \xbr_{i}:i \xbe I\}}{card(I)},$
$ \xbr_{i}$ the reliability in dimension $i.$
 \xDH
We may want to give more reliable messages etc. more weight.
E.g. for $ \xbr,$ $ \xbr',$ we may want to calculate the weighted mean
of $ \xbr,$ $ \xbr',$
depending on their individual reliabilities.
We can do this as follows:

Consider $av(\xbr)$ and $av(\xbr').$ If $av(\xbr) \xcg av(\xbr'
),$ take
$d:=$ $ \frac{1}{2}+ \frac{1}{4}(av(\xbr)-av(\xbr')).$ Then $d \xbe
[0.5,1],$ and this will be the
weight for $ \xbr.$ The weight for $ \xbr' $ will be $1- \xCf d.$ If
$av(\xbr)<av(\xbr'),$ similarly.
 \xEj
\subsection{
Combinations
}

{\tiny LABEL: {Section Combinations}} \\[1mm]
\label{Section Combinations}

\ed

We discuss now a number of cases.
The solutions are suggestions, often, one will find alternatives which
might be as good or even better. Our discussion is more centered on basic
ideas
than on details - which might depend on context, too.
A good overall algorithm will probably be quite robust against local
changes.

In the following, we will treat only conflicts between two
agents/messages.
Of course, also situations like three values, 8, 9, and -1, need to be
treated
(example due to D. Makinson), where we will give more credibility to
$\{8,9\}$
than to the exceptional value -1. We treat in the following only pairs, so
$\{8,9\},$ $\{-1,8\},$ $\{-1,9\},$ which should go in the same sense as
treating the
triple $\{-1,8,9\}.$
\subsubsection{
$\xbr(A)$ and $\xbr(M)$
}

{\tiny LABEL: {Section A-M}} \\[1mm]
\label{Section A-M}

 \xEh
 \xDH

From $ \xbr (A)$ to $ \xbr (M):$

If agent $ \xCf A$ sends message $M,$ without any reliability, the
reliability of $M$ will by default be
the reliability of $ \xCf A.$ (It may be adjusted later, due to other
messages.)
If it has already an initial reliability, the reliability
will be the combined (in the spirit of AND) reliability of the agent and
the
message.

 \xDH

From $ \xbr (M)$ to $ \xbr (A):$

When $ \xbr (M)$ was modified, this should have repercussions on $ \xbr
(A).$
If $ \xbr (M)$ was increased, $ \xbr (A)$ should increase, too,
if $ \xbr (M)$ was decreased, $ \xbr (A)$ should decrease, too.
The effect should be ``dampened'' however. One wrong message should not
totally destroy the reliability of the agent.
For this purpose, we have introduced the inertia (of the agent). The
bigger the
inertia,
the less we change $ \xbr (A).$ E.g., inertia may code the number of
values which
went already into $ \xbr (A).$

We may do this with a new message going ``backwards''. If we store the path
of the message, we avoid going in cycles, i.e. going again to $ \xbr (M),$
etc.

 \xEj
\subsubsection{
Chains of Messages and Reliabilities
}

{\tiny LABEL: {Section Chains}} \\[1mm]
\label{Section Chains}

See Example \ref{Example Temp} (page \pageref{Example Temp}), Case 3.

Suppose agent $ \xCf A$ sends message $M,$ and agent $A' $ passes $M$ on,
perhaps
with some modification, so this is message $M'.$
 \xEh
 \xDH
The combined message $ \xCf MM' $ will have some $ \xbr (MM').$ It seems
natural to
set $ \xbr (MM')$ $ \xck $ $min(\xbr (M), \xbr (M')).$
 \xDH
Conversely: Suppose we have modified $ \xbr (MM'),$ and given it a new
value $ \xbr' (MM').$
We have to break down the modification to new $ \xbr' (M)$ and $ \xbr'
(M'),$ in a
reasonable way, so that again $ \xbr' (MM')$ $ \xck $ $min(\xbr' (M),
\xbr' (M')).$

Recall that the old $ \xbr (MM')$ was calculated as $ \xck min(\xbr (M),
\xbr (M')),$
say $ \xbr (MM')= \xba *min(\xbr (M), \xbr (M')),$ where $0 \xck \xba
\xck 1.$
We adjust now $ \xbr (M)$ and $ \xbr (M')$ to $ \xbr' (M)$ and $ \xbr'
(M')$ such that again
$ \xbr' (MM')= \xba *min(\xbr' (M), \xbr' (M')),$ e.g. using the
same factor on $ \xbr (M)$ and
$ \xbr (M').$
 \xEj
\subsubsection{
Two Parallel Messages
}

{\tiny LABEL: {Section Parallel}} \\[1mm]
\label{Section Parallel}

Different agents might send messages with reliability $ \xbr $ and $ \xbr
' $
respectively about the same subject. Those messages
might agree, or not.

Case 1:

The messages agree. The reliabilities support each other, and both $ \xbr
$ and
$ \xbr' $ should increase, where a smaller one should perhaps increase
more.

Case 2:

The messages disagree. The reliabilities contradict each other, and both $
\xbr $ and
$ \xbr' $ should decrease, where a smaller one should perhaps decrease
more.

Details are left to the reader.
\subsubsection{
Two Parallel Messages About Reliability
}

{\tiny LABEL: {Section Par-Rel}} \\[1mm]
\label{Section Par-Rel}

Suppose agent $ \xCf A$ sends message $M$ with reliability $ \xbr,$ $A' $
$M' $ with
reliability $ \xbr',$ the contents of $M$ is a reliability $ \xbm,$
that of
$M' $ a reliability $ \xbm',$ where $ \xbm $ and $ \xbm' $ are about
the same agent $A'' $ or
message $M''.$ Suppose that $A'' $ or $M'' $ has old reliability $ \xbn
.$ We want to calculate a
new reliability $ \xbn' $ (of $A'' $ or $M'').$

It seems reasonable to calculate the new reliability $ \xbn' $ from
$ \xbm,$ $ \xbm',$ $ \xbn,$ respecting inertia $ \xbi $ of $ \xbn,$
and the respective reliabilities
$ \xbr $ and $ \xbr'.$ The new inertia of $ \xbn' $ should increase, as
more values
go into $ \xbn' $ than went into $ \xbn.$

Details are left to the reader.
\section{
Discussion
}

Our approach is very pragmatic,
and takes its intuition from e.g. physics, where a theory is
considered true - but revisably so! - when there is ``sufficient''
confirmation, by experiments, support from other theories, etc.

Many human efforts are about establishing reliability of humans or
devices. An egineer or physician has to undergo exams to assure that
he is competent, a bridge has to meet construction standards, etc.
All this is not infallible, experts make mistakes, new, unknown
possibilities of failure may appear - we just try to do our best.

Our ideas in Section 
\ref{Section Details} (page 
\pageref{Section Details})  are examples how it can be
done,
but no definite solutions. The exact choice is perhaps not so
important, as long as there is a process of permanent adjustment.
This process has proven extremely fruitful in science, and deserves
to be seen as a powerful method, if not to find truth, at least to find
``sufficient'' information.

From an epistemological point of view, our position is that of
``naturalistic epistemology'', and we need not decide between
``foundationalism'' and ``coherentism'', the interval $[-1,1]$ has
enough space to maneuvre between more and less foundational
information. See e.g.  \cite{Sta17c}.

Our approach has some similarities with the utility approach,
see the discussion in  \cite{BB11}, the chapter on utility.
An assumption, though false, can be useful: if you think a lion is
outside,
and keep the door closed, this is useful, even if, in fact, it is a tiger
which is outside. ``A lion is outside'' is false, but sufficiently true.
We think that this shows again that truth should not be seen as something
absolute, but as something we can at best approximate; and, conversely,
that it is not necessary to know ``absolute truth''.
We go beyond utility, as improvement is implicit in our approach.
Of course, approximation may only be an illusion generated by the fact
that we develop theories which seem to fit better and better, but whether
we approach reality and truth, or, on the contrary, move away from reality
and
truth, we cannot know.

There are many things we did not consider, e.g. if
more complicated, strongly connected, structures have stronger inertia
against
adjustment.

We have an
example structure which handles these problems very well: our brain.
Attacks, negative values of reliability, correspond to inhibitory
synapses,
positive values, support, to excitatory synapses. Complex, connected
structures
with loops are created all the time without uncontrolled feedback.
It is perhaps not sufficiently clear how this works, but it must work!
(The ``matching inhibition'' mechanism seems to be a candidate. See also
 \cite{OL09} for a discussion of the ``cooperation'' of excitatory and
inhibitory inputs of a neuron. E.g., excitation may be followed closely by
inhibition, thus explaining the suppression of such feedback.
The author is indebted to Ch. von der Malsburg, FIAS, for these hints.)
Our theories about the
world survive some attack (inertia), until ``enough is enough'', and we
switch
emphasis. The brain's mechanisms for attention can handle this.
\subsection{
Is this a Theory of Truth?
}

The author thinks that, yes, though
we hardly mentioned truth in the text.

Modern physics are perhaps the best attempt to find out what
``reality'' is, what ``truly holds''. We had the development of physics in
mind,
reliability of experiments, measurements, coherence of theories (forward
and backward influence of reliabilities), reputation of certain
physicists,
predictions, etc. Of course, the present text is only a very rough sketch,
we see it as a first attempt, providing some highly flexible ingredients
for a
more complete theory in this spirit.
\section{
Acknowledgements
}

The author would like to thank David Makinson for very helpful comments.

\end{document}